\documentclass[runningheads]{llncs}
\usepackage{graphicx}
\usepackage{mathptmx}
\usepackage{amsmath}
\usepackage{soul}
\usepackage{color}
\usepackage[dvipsnames]{xcolor}
\usepackage{amsfonts}
\usepackage{amssymb}
\usepackage{verbatim}
\begin{document}
\title{Comparing vector fields across surfaces: interest for characterizing the orientations of cortical folds}  
\titlerunning{Comparing vector fields across surfaces}
%
\author{Amine Bohi\inst{1} \orcidID{0000-0002-2435-3017} \and Guillaume Auzias\inst{2} \orcidID{0000-0002-0414-5691} \and Julien Lefèvre\inst{2} \orcidID{0000-0003-3670-6112}}
\authorrunning{A. Bohi et al.}
%
\institute{Laboratoire d'Informatique et Systèmes, Aix-Marseille Université, CNRS UMR 7020. Marseille, France \\ Institut de Neurosciences de la Timone, Aix-Marseille Université, CNRS UMR 7289. Marseille, France}

%
\maketitle              
\begin{abstract}

Vectors fields defined on surfaces constitute relevant and useful representations but are rarely used. One reason might be that comparing vector fields across two surfaces of the same genus is not trivial: it requires to transport the vector fields from the original surfaces onto a common domain. In this paper, we propose a framework to achieve this task by mapping the vector fields onto a common space, using some notions of differential geometry. The proposed framework enables the computation of statistics on vector fields.
We demonstrate its interest in practice with an application on real data with a quantitative assessment of the reproducibility of curvature directions that describe the complex geometry of cortical folding patterns. 
The proposed framework is general and can be applied to different types of vector fields and surfaces, allowing for a large number of high potential applications in medical imaging.


\end{abstract}
\section{Introduction}

In the neuroimaging community the diversity of acquisition modalities and image processing techniques has led to the emergence of various mathematical objects serving as formal representations of specific features of the structure and functioning of the brain. Vector fields have been commonly used to represent tensors, ODFs or fibers derived from Diffusion MRI. They have also been used in the context of tensor-based morphometry to characterize changes from one brain anatomy to another or to a template \cite{ashburner1998identifying}. For example in \cite{hadj2016longitudinal} the authors estimated longitudinal brain deformations represented as stationary vector field at the individual level and mapped them in a common space through parallel transport. In all these examples, the vector fields are computed and analyzed in the 3D Euclidian space.

Vector fields can also be computed on a mesh to represent data lying on the 2D manifold corresponding to a cortical surface. Such representations have been used to track the local propagation of neural activities \cite{lefevre2009optical}, to describe quantitatively the directions of cortical folds \cite{boucher2009oriented} or to offer a visualization tool for connectivity analysis \cite{bottger2014connexel}. 
\cite{li_automatic_2009} introduced a method to parcellate the cortical surface into sulcal regions based on vector fields representing principal curvature directions.
More specifically, the authors of \cite{boucher2009oriented} proposed a method to estimate a dense orientation field on a surface in the direction of its folds. The method is based on Tikhonov regularization of principal curvatures that works directly on a triangulated mesh and was validated on synthetic data. They applied it to identify direction-specific changes in the shape of cortical surfaces from healthy subjects and patients with Alzheimer's disease. 
To our knowledge, this is the only work reporting an analysis of vector fields that are defined on the cortical surface and compared across individuals (and thus across different geometries).
However even in that paper, the comparison across individuals is done through a scalar metric and not \textit{directly on the vector fields}.

\begin{figure}
    \centering
    \includegraphics[width=\linewidth]{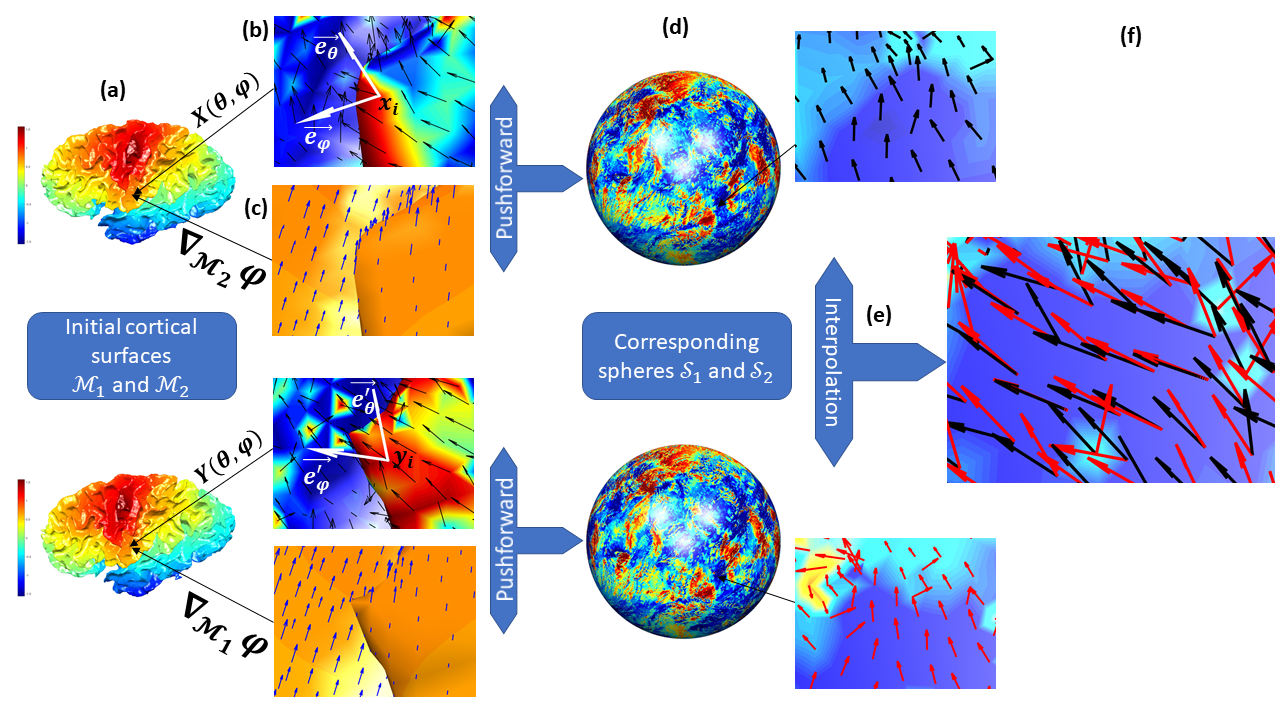}
    \caption{Our framework to compare the vector fields from two different cortical surfaces, detailed in section \ref{methods}. From left to right : (a) the two cortical surfaces $\mathcal{M}_1$ and $\mathcal{M}_2$. (b) The vector fields defined on $\mathcal{M}_1$ and $\mathcal{M}_2$ resp. (e.g. curvature principal directions). (c) Gradients coefficient calculated on $\mathcal{M}_1$ and $\mathcal{M}_2$. (d) Pushforward of each vector field on the corresponding sphere $\mathbb{S}_1$ and $\mathbb{S}_2$. (e) Resampling of the vector fields from each sphere onto a common domain. (f) The two vector fields can then be quantitatively compared at each location of the common domain. The color-coded information in (a) and (c) corresponds to one of the spherical coordinate fields (see section\ref{implementation}). In (b), (d) and (f) it shows the coordinate of the vector field along the vector $\mathbf{e}_{\varphi}$ of the local basis (white arrows in (b)).}
    \label{fig:schema}
\end{figure}

In this work we propose a methodological framework illustrated on Fig.\ref{fig:schema} allowing to transport a vector field from one cortical surface with spherical topology to the sphere, which is mandatory for comparing vector fields across cortical surfaces from different individuals.
The proposed framework thus enables the computation of \textit{statistics on vector fields}. 

In section 2, we recall the notions of differential geometry we used and show how we can compute easily the pushforward operator between two submanifolds. In section 3, we demonstrate the interest of our approach in an exemplar application to real data.
We compare vector fields representing the orientation of principal curvature directions in a scan-rescan context, allowing to quantitatively assess the reproducibility of the estimation of principal curvature directions.
A specific emphasis is put on principal curvature directions based on previous works showing promising applications of this particular type of vector fields \cite{li_automatic_2009,boucher2009oriented}, but our framework can be applied to any kind of vector fields defined on triangulated surfaces with a spherical topology.

\section{Method}
\label{methods}
\subsection{Mathematical background}
\label{mathback}
Before detailing our approach for transferring information vectors from one surface to another, we first recall general differential geometry background. In the following, vectors are represented in bold, so there is no ambiguity for instance between $\mathbf{n}$ a normal vector and \textit{n} an integer value.

\paragraph{Submanifold}

We define $\mathcal{M}$ a submanifold of $\mathbb{R}^{n+1}$ (in practice $n = 2$) as a subset of $\mathbb{R}^{n+1}$ characterized by a finite union of open sets $U_i$ and diffeomorphic maps $\varphi_{i} : U_{i} \rightarrow \mathbb{R}^n$. $(U_{i},\varphi_{i})$ is often called local charts or sometimes parameterization domain in neuroimaging. In the following we will simply denote the map $\varphi : p \rightarrow (x_1(p),...,x_n(p))$\\
It allows defining a tangent space $T_{p} \mathcal{M}$ at a point $p$ that is spanned by the column vectors of $\mathbb{R}^n$, $\mathbf{e}_i=\partial_{x_i} \varphi$ that are the variations along the surface when each axis of the parameterization domain varies. $\mathbf{e}_i$ depends on the local geometry around each point $p$. Alternatively the tangent space is orthogonal to the normal vector $\mathbf{n}$ whose expression is simple when $n=2$ (cross product).\\

\paragraph{Vector field}
A vector field $X$ is a map  which at any point $p$ of $\mathcal{M}$ associates a vector $\mathbf{X}(p)$ in the tangent space $T_{p}M$. Locally, at each point $p$ it can be written as:
$$ \mathbf{X}(p) = \sum_{i=1}^n X_i(p) \mathbf{e_i} $$ 
$X_i$ are the coordinates of the vector field in the local basis.

\paragraph{Differential}

Given a smooth map $f : \mathcal{M} \rightarrow \mathbb{R}$ we define the differential of $f$ at $p$ as the linear application denoted $D_{p} f$ :
$$ \mathbf{v} \in T_{p} \mathcal{M} \rightarrow \sum_{i=1}^n \frac{\partial (f \circ \varphi^{-1})}{\partial x_i}(p) v_i \in \mathbb{R}$$
where $v_i$ are the coordinate of the input vector. $f \circ \varphi^{-1}$ allows to come back in the local charts where the partial derivatives are computed.

\paragraph{Vector field under a diffeomorphism} 
Considering $F : \mathcal{M} \rightarrow \mathcal{N}$ a diffeomorphism between two submanifolds and $X$ a vector field on $\mathcal{M}$. The corresponding vector field on $\mathcal{N}$ denoted $F_*\mathbf{X}$ is defined canonically by:

\begin{equation*}
    F_*\mathbf{X} : q \rightarrow \big(D_{F^{-1}(q)} F\big) (\mathbf{X}(F^{-1}(q)))
\end{equation*}
This operation is called \textit{pushforward} of $X$.\\
Note that it involves the differential of $F$ on the image point of $q$ by the inverse diffeomorphism, applied to the corresponding vector of the tangent plane.\\
This formula is nicely interpretable but cannot be implemented directly in a discrete setting. We therefore examine another strategy involving the gradient operator.

\subsection{Gradient trick}
\label{gradient}

We recall two definitions before introducing our implementation of the pushforward of a vector field.

\paragraph{Metric}
$\mathcal{M}$ is equipped by a metric $g$ if there is locally an application from $\mathcal{M} \times \mathcal{M} $ in $\mathbb{R}$. The matrix $G = (g(\mathbf{e}_i,\mathbf{e}_j) )$ encodes all the local geometric information and is called first fundamental form.

\paragraph{Gradient} 
Let $f$ be a smooth map $f : \mathcal{M} \rightarrow \mathbb{R}$.
 the gradient of $f$ at $p$ is defined as the vector field such as at each $p$:
$$ g\big(\nabla f(p), \mathbf{v} \big) = (D_p f) \mathbf{v}, $$ 
which is valid for every vector $\mathbf{v}$ and in particular those of the local basis.
The following lemma is a direct consequence of this definition, keeping in mind that $D_p x_i$ is the linear form that maps any vector $\mathbf{v}$ to its $i^{th}$ coordinate.

\begin{lemma}
We can easily compute $\nabla x_i$ and we have the change of basis formula
\begin{equation}\nabla x_i = \sum_{i=1}^n \alpha_{i,j} \mathbf{e}_j  \hspace{1cm} \mathbf{e}_i = \sum_{i=1}^n \beta_{i,j} \nabla x_j \label{change:basis} \end{equation}
where for a fixed value of $i$, the vector $(\alpha_{i,j})_j$ (resp. $(\beta_{i,j})_j$) equals the $i^{th}$ column of $G^{-1}$ (resp. $G$). Moreover the matrix $(g(\nabla x_i,\nabla x_j))$ equals $G ^{-1}$.
\end{lemma}

Now we stress a trivial but relevant observation:
given $\mathcal{M}$ and $\mathcal{N}$ two submanifolds with local coordinates $(x_i)$ and $(y_i)$ respectively and a diffeomorphism $F: \mathcal{M} \rightarrow \mathcal{N}$ such as its differential is identity everywhere, if $\mathbf{X}$ is a vector field on $\mathcal{M}$, one can read:
$ \mathbf{X}(p) = \sum_{i=1}^n X_i \mathbf{e}_i  $ and the pushforward is obtained as 
$F_* \mathbf{X}(q) = \sum_{i=1}^n X_i F_*\mathbf{e}_i = \sum_{i=1}^n X_i \mathbf{f}_i$ where $\mathbf{f}_i$ is the canonical basis of the tangent plane.

\begin{lemma}[Gradient trick]
Given the previous lemma, the vector field and its pushforward can be decomposed as 
$$ \mathbf{X}(p) = \sum_{i=1}^n X_i \mathbf{e}_i  \hspace{1cm} F*\mathbf{X}(p) = \sum_{i=1}^n Y'_i \nabla y_i  $$
where, stating that $X = (X_1,...,X_n)^{\top}$ (the same for $Y'$):
$ Y' = \tilde{G} X $
and $\tilde{G}$ follows the same convention as in Lemma 1 but for $y_i$ coordinates. Moreover $X_i = g(\mathbf{X}(p),\nabla x_i)$
\end{lemma}
The proof is straightforward by using first that $g(\mathbf{X}(p),\nabla x_i) = \sum_{j} X_j g(\mathbf{e}_j,\nabla x_i) $ and $g(\mathbf{e}_j,\nabla x_i)$ vanishes unless $i=j$. Second the expression of $F*\mathbf{X}(p)$ is a consequence of the change of basis of lemma 1 applied to the previous observation when the pushforward is identity.

The main advantage of this lemma lies in the decomposition of the vector field in the basis formed by the gradient of the coordinates.
Indeed, the gradient operator can be computed at each vertex of a discrete surface by using the finite elements method as in \cite{lefevre2008optical}. 

\subsection{Implementation aspects}
\label{implementation}
Under the assumptions on the diffeomorphism between the two surfaces, the previous framework is very general and does not require any condition on the topology of the two submanifolds. In our application we will work on cortical surfaces represented as discrete triangular meshes with a spherical topology for which a spherical parameterization can been obtained for instance as described in \cite{Fischl1999a}. This gives us the spherical coordinates  $(x_1,x_2) = (\theta,\varphi)$ of each point of a cortical mesh $\mathcal{M}$. This mapping defines a diffeomorphism from the cortical surface onto the sphere canonically defined with same $(\theta,\varphi)$ coordinates  whose pushforward is equal to identity. Lemma 2 can be applied and it yields the following steps to obtain the pushforward of a vector field $\mathbf{X}$  onto the sphere:
\begin{enumerate}
    \item Compute $\nabla_{\mathcal{M}} x_i$ (resp$\nabla_{\mathcal{\mathbb{S}}^2} y_i$) on the cortical mesh (resp  on the spherical mesh) 
    \item Compute $\tilde{G}^{-1}= (g(\nabla y_i,\nabla y_j)) $  given $g$ is the inner product of $\mathbb{R}^3$
    \item Compute $X=(g(\mathbf{X}, \nabla x_i))$ and apply $\tilde{G}$ to get $Y'$
\end{enumerate}

Once the vector fields are transferred from each cortical surfaces onto the corresponding spheres, we interpolate all the vector fields from every spheres on a unique isocahedron spherical mesh (160k vertices).
We chose to use nearest-neighbor interpolation and not linear or higher order interpolation technique here in order to avoid the smoothing effect that would be induced.

\section{Experiments and Results}
\label{experiments}

\subsection{Data and Preprocessings}
We use the publicly available dataset Kirby21 (https://www.nitrc.org/projects/multimodal/, \cite{Landman2011}) consisting of acquired scan-rescan imaging sessions on 21 healthy volunteers (11 M/10 F, 22–61 years old).
The short delay between the two acquisitions from each subject allows to assess the reproducibility of image processing techniques by limiting as much as possible the potential biological changes across the two scans.
We apply our framework to compare between the two scans of each subject the orientation of the principal directions of the curvature estimated in each point of the highly folded cortical surfaces, hereby allowing to assess the reproducibility of the estimation of the principal directions (see section\ref{curvature}).


For each scan and rescan MRI data from each subject (2 x 21 = 42), we extracted the triangulated cortical surfaces for each hemisphere (left and right) using the well validated freesurfer pipeline \cite{Fischl1999a}.
Freesurfer segmentation and surface extraction have been shown to be highly reproducible, e.g. \cite{Fjell2009}, which is mandatory to assess the reproducibility of principal direction estimations properly.
In this work we used the 'white' surface that corresponds anatomically to the boundary separating gray and white-matter tissues.
This surface corresponds to a well-marked intensity gradient in the original MRI and is easier to properly extract than the 'pial' surface for which accurate distinction between cortical tissue and dura-matter can be problematic.
All surfaces were visually inspected for potential segmentation inaccuracy, no manual correction was required for the 'white' surface.

\subsection{Curvature directions}
\label{curvature}

In differential geometry, curvature tensor at a point of a submanifold can be estimated by computing second order derivatives with respect to the local coordinates. When $n=2$ it allows finding a local coordinate system in which a Taylor expansion yields a simple relation between the 3d coordinates
$z(x,y)=\frac{1}{2}\kappa_m x^2 + \frac{1}{2}\kappa_M y^2  $.
$\kappa_m$  and $\kappa_M$ are the principal curvatures and they are associated to orthogonal directions \cite{rusinkiewicz2004estimating}. 
Several approximation methods have been proposed to compute mean curvature $\frac{1}{2}(\kappa_m + \kappa_M)$ and gaussian curvature $\frac{1}{2}(\kappa_m + \kappa_M)$ on discrete meshes \cite{vavsa2016mesh}, and some of these estimate explicitly the curvature directions. In the present work, we use the method \cite{rusinkiewicz2004estimating} based on a finite difference scheme.


\subsection{Reproducibility at the individual level}

\begin{figure}
    \centering
    \includegraphics[width=0.48\linewidth]{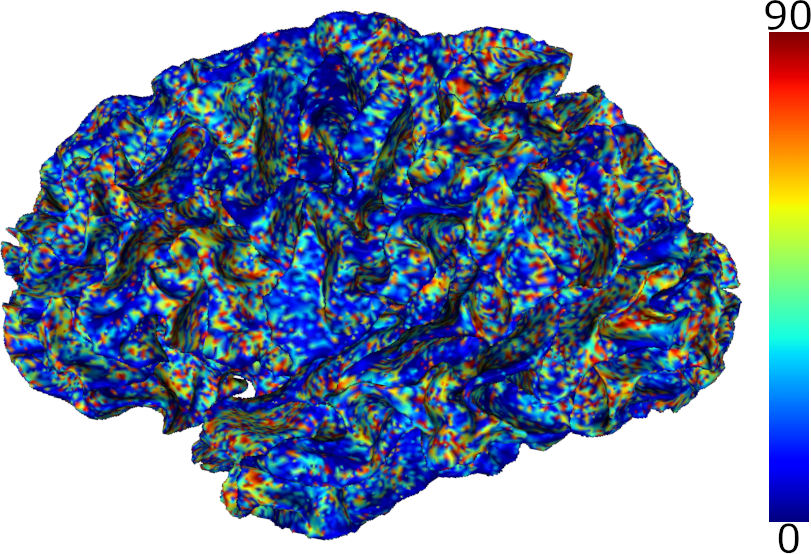}
    \includegraphics[width=0.5\linewidth]{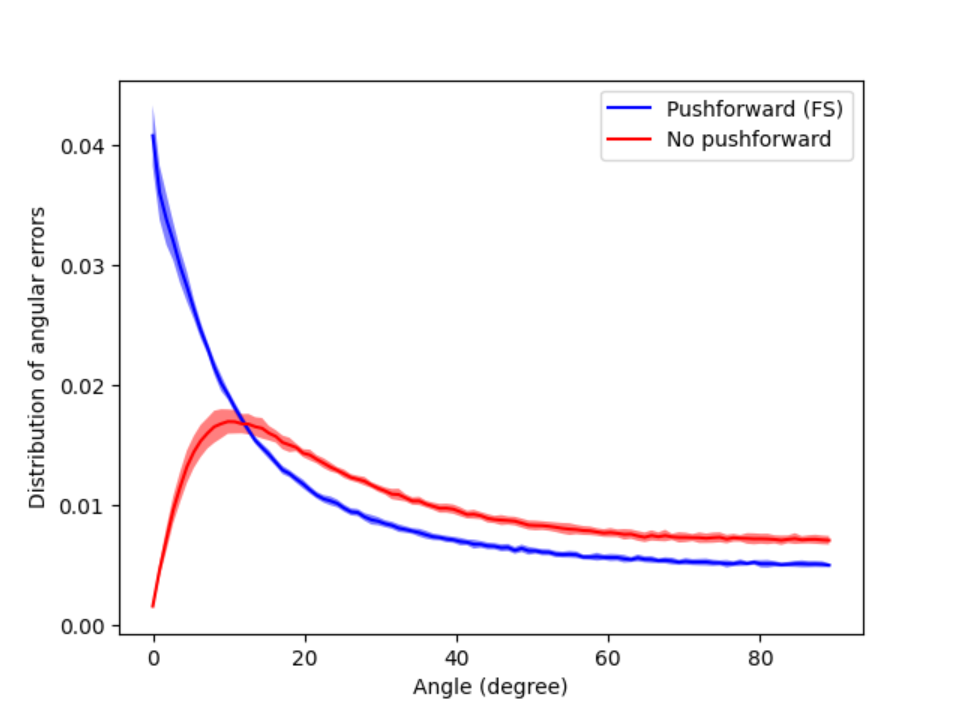}
    \caption{Left: Map showing the angular error between curvature directions of the scan and rescan in each vertex of the cortical surface of an exemplar subject. Right: Average across the 21 individuals of the distribution of angular errors using our framework in blue and without in red.}
    \label{fig1}
\end{figure}

First we applied our framework at the individual level to compute the angles between the curvature directions estimated on the surfaces corresponding to scan and rescan respectively. An example of angular errors map for an individual is depicted on Figure \ref{fig1}, with the distribution of the angular errors across every vertices and across the 21 individuals of the group. For comparison with a naive approach, we also computed the angular errors between the original vector fields, i.e. without using the pushforward. The major difference between the two distributions demonstrates clearly that the naive approach is inappropriate, with a peak in the distribution around 15 degrees which would suggest a clear lack in the reproducibility of the estimated curvature directions.
With the pushforward, the angular errors are much lower, inferior to 20 degrees for the majority of cortical locations.


\subsection{Group analysis of curvature directions}

\begin{figure}
    \centering
    \includegraphics[width=0.48\linewidth]{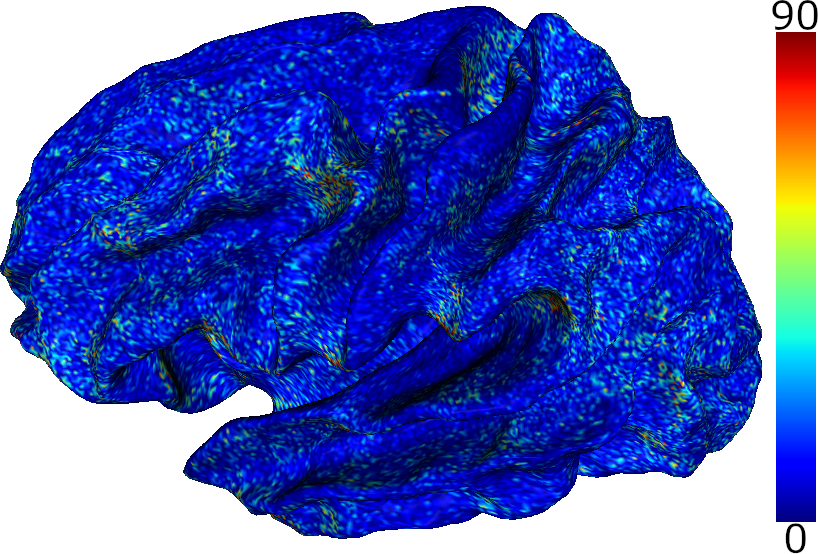}
    \includegraphics[width=0.5\linewidth]{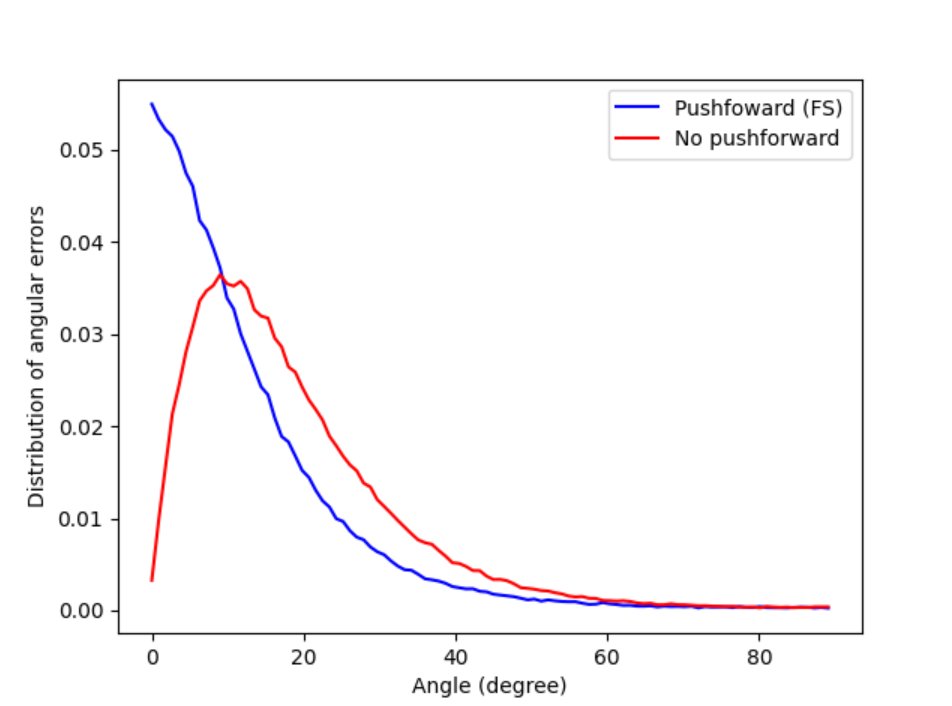}
    \caption{Left: Angle between the average vector fields for test and retest obtained with pushforward. Right: Distribution of angles between the average vector fields obtained with and without pushforward.}
    \label{fig2}
\end{figure}

We then applied our approach to compute the average across individuals of the vector fields for each acquisition (scan and rescan resp.), and computed the angular error between the average vector fields. 
On Fig \ref{fig2} left we show the map of the angular errors on an average surface computed using our framework. On the right part we show the distributions of angular errors with and without pushforward. Only the errors computed using our framework really reflect the reproducibility of the estimated principal directions of curvature.
Thanks to the framework introduced in present work, we will be able to assess and compare the reproducibility of different estimation techniques in a follow-up study.

\section{Conclusion}
\label{conclusion}

The methodological framework we introduced allows to transfer any kind of vector field from folded surfaces onto a common sphere, which was the lacking step for running quantitative comparisons and group studies. 
Beyond the curvature directions considered in the present work, a large number of different neuroimaging applications might be of major interest, such as analysing surfacic gradients of functional connectivity obtained from resting state fMRI or deformation fields resulting from surface-registration techniques.
Other types of surfaces could also be considered, such as cerebellar reconstructions or other structures such as the hippocampus or any other deep cerebral structures.
More generally, our framework could be applied to surfacic representation of any organ such as kidney or prostate in group studies such as case-control comparison, thus opening a large number of high potential applications in medical imaging.

%
%

%
%
%
\bibliographystyle{splncs04}
\bibliography{ref}
%


\end{document}